\documentclass[letterpaper]{article}
\usepackage{aaai_paper/aaai2026}
\usepackage{times}
\usepackage{helvet}
\usepackage{courier}
\usepackage[hyphens]{url}
\usepackage{graphicx}
\usepackage{natbib}
\usepackage{caption}
\usepackage{amsmath}
\usepackage{amssymb}
\usepackage{algorithm}
\usepackage{algorithmic}
\usepackage{newfloat}
\usepackage{listings}% 以下是自己加的package--------------------------------------------------------
\usepackage{booktabs}    % for \toprule, \cmidrule, \bottomrule
\usepackage{multirow}
\usepackage{mathtools}
\usepackage{float}
\usepackage{pifont}
\usepackage{subcaption}

\newcommand{\cmark}{\ding{51}}   % check mark
\newcommand{\xmark}{\ding{55}}   % cross mark

\DeclareCaptionStyle{ruled}{labelfont=normalfont,labelsep=colon,strut=off}
\floatstyle{ruled}
\newfloat{listing}{tb}{lst}{}
\floatname{listing}{Listing}

\title{Your Paper Title}
\author{
Zhehong Ren \thanks{Equal contribution} ,
Tianluo Zhang\footnotemark[1],
Yiheng Lu\footnotemark[1],
Yushen Liang\footnotemark[1],
Promethee Spathis
}
\affiliations{
New York University Shanghai\\
\{zr2245,tz2668,yl12003,yl11949,ps147\}@nyu.edu
}
\title{Relation-Aware LNN–Transformer for Intersection-Centric Next-Step Prediction}
\begin{document}
\maketitle
% \section*{Abstract}

% TODO: Write Abstract here.
\begin{abstract}
Next-step location prediction plays a pivotal role in modeling human mobility, underpinning applications from personalized navigation to strategic urban planning. However, approaches that assume a closed world—restricting choices to a predefined set of points of interest (POIs)—often fail to capture exploratory or target-agnostic behavior and the topological constraints of urban road networks. Hence, we introduce a road-node-centric framework that represents road-user trajectories on the city’s road-intersection graph, thereby relaxing the closed-world constraint and supporting next-step forecasting beyond fixed POI sets. To encode environmental context, we introduce a sector-wise directional POI aggregation that produces compact features capturing distance, bearing, density, and presence cues. By combining these cues with structural graph embeddings, we obtain semantically grounded node representations. For sequence modeling, we integrate a Relation-Aware LNN–Transformer—a hybrid of a Continuous-time Forgetting Cell (CfC-LNN) and a bearing-biased self-attention module—to capture both fine-grained temporal dynamics and long-range spatial dependencies. Evaluated on city-scale road-user trajectories, our model outperforms six state-of-the-art baselines by up to 17 percentage points in Acc@1 and 10 percentage points in MRR, and maintains high resilience under noise—losing only 2.4 percentage points Acc@1 at 50\,m GPS perturbation and 8.9 percentage points Acc@1 at 25\% POI noise.
\end{abstract}

% Our contributions are: (1) a novel open-world formulation over road graphs; (2) POI-region projection for topology-semantic fusion; (3) a hybrid temporal-graph encoder; and (4) comprehensive evaluation on accuracy, robustness, and efficiency.
\section{Introduction}

Predicting the next intersection for a road user—across walking, cycling, and driving—is a cornerstone of human mobility modeling, driving applications in personalized navigation, traffic forecasting, and urban planning. A large body of next‑POI work frames prediction as sequence modeling over a limited POI vocabulary, often overlooking (i) the \emph{topological constraints} of the road graph and (ii) the \emph{directional distribution} of nearby semantics. For instance, Transformer‑based approaches typically treat trajectories as flat token streams without encoding intersections as decision points or road connectivity, which can induce candidate bias and a coarse treatment of street structure (e.g.~\cite{TrajTrans}). Even graph‑enhanced POI recommenders such as GETNext~\cite{yang2023getnext} still operate on POI tokens rather than directly on the road‑intersection state space.

Nevertheless, empirical road‑user movement exhibits substantial exploration: using a high‑precision, longitudinal dataset of 400+ users, Cuttone et~al.\ report that \emph{20–25\%} of transitions lead to previously unseen places and that roughly \emph{70\%} of locations are visited only once~\cite{CuttoneLehmannGonzalez2016}. Such exploratory behavior directly challenges fixed‑vocabulary next‑POI predictors and motivates modeling unbounded by a predefined POI token set.

We address such gaps with a \emph{Road-Node–Centric Next-Step Prediction} framework that represents trajectories on the city’s intersection graph and combines static semantic context, topology, and geography information. The three main contributions of our paper are the following:
\begin{itemize}
  \item \textbf{Sector‑wise, direction‑aware node representation.}
We partition each intersection’s neighborhood into equal‑angle sectors and, within each sector, summarize multi‑category POI cues (distance, bearing, density, presence) to form a compact, inductive node descriptor. This sectorization integrates directionality at the feature level—capturing the anisotropy of urban space and aligning with outgoing road bearings—so that one‑hop candidates that are topologically similar but differ by direction become distinguishable.

  \item \textbf{Relation-Aware LNN–Transformer for spatiotemporal encoding.}
  We build a unified architecture by first encoding trajectory dynamics with a Continuous-time Forgetting Cell (CfC-LNN) and then applying a Relation-Aware Transformer that adds a bearing-based bias to its self-attention logits. The Rel-Aware LNN–Transformer captures fine-grained temporal changes, and the bearing-biased attention captures long-range directional dependencies on the road graph—all within a compact 2.34M-parameter model.

  \item \textbf{Empirical and robustness gains.}
  On city-scale road-user trajectories (1--10\,km), our model surpasses six competitive baselines in Accuracy@1 and MRR, and degrades gracefully under realistic GPS perturbations and POI-feature noise. We attribute the robustness partly to the stable continuous-time recurrence in CfC-LNN (closed-form state updates) coupled with topology-aware attention; training remains gradient-based, but the forward dynamics improve conditioning and sample efficiency.
  \end{itemize}

%While our results focus on a single city‑scale dataset, the framework is designed to relax closed‑world assumptions by operating on intersections rather than fixed POI vocabularies. Broader cross‑city validation and larger candidate sets are left to future work.

%\begin{figure*}[!t]
%  \centering
%  \includegraphics[width=0.4\textwidth]{pics/flowchart.pdf}
%  \caption{\textbf{Overview of our Relation‑Aware LNN–Transformer.}
%  A road‑user trajectory is projected onto the intersection graph and encoded via sector‑wise POI and geometric features fused with Node2Vec embeddings.  
%  The differenced features go through a CfC‑LNN and $L\!-\!1$ standard Transformer layers, followed by a relation‑aware attention block that biases self‑attention with inter‑node bearings, and finally a softmax over one‑hop candidates.}
%  \label{fig:overview}
%\end{figure*}
\section{Related Work}

\subsection{Liquid Neural Networks (LNNs)}
Liquid neural networks model sequence dynamics in continuous time with input‑modulated state equations, offering compact models and strong stability/robustness properties relative to ODE/RNN baselines. 
Early ``Liquid Time‑constant'' (LTC) models define linear first‑order dynamics with gates and are trained by backpropagation through time (BPTT)~\cite{hasani2020ltc}. 
Closed‑form continuous‑time (CfC) variants further avoid heavy numerical solvers by deriving analytic updates, improving efficiency while preserving expressivity~\cite{hasani2022cfc}. 
LNNs have demonstrated resilience to distribution shifts in control and navigation, e.g., robust out‑of‑distribution flight with vision inputs~\cite{hasani2023scirobotics}. 
These properties motivate our use of a CfC‑LNN encoder for irregular dwell times and non‑uniform sampling in trajectories.

\subsection{Next POI/Location Prediction}
\paragraph{Sequence‑only models.} 
RNN variants inject temporal and spatial intervals into gates or transitions but treat POIs as vocabulary tokens without explicit road topology: ST‑LSTM with time/distance gates~\cite{zhao2018stlstm}, STRNN with time/distance‑specific transition matrices~\cite{liu2016strnn}, DeepMove with historical attention~\cite{feng2018deepmove}, and ARNN with attentional sequence modeling~\cite{guo2020arnn}.  

\paragraph{Transformer‑based models.} 
Recent Transformers capture long‑range dependencies yet typically operate on flat token sequences; examples include GETNext, which augments a Transformer with a global trajectory flow map for next‑POI recommendation~\cite{yang2023getnext}, and STTF‑Recommender for mining non‑contiguous visit relations~\cite{TrajTrans}.  
Although GETNext and STTFRecommender both advance the next POI prediction, neither incorporates explicit road-graph connectivity or direction-aware POI semantics, which our intersection-level sector-wise aggregation is designed to remedy.
\subsection{Input Embeddings and Information Aggregation}
\paragraph{Structural priors.} Node2Vec learns transductive structural embeddings via biased random walks that mix homophily and structural equivalence~\cite{grover2016node2vec}. 
GraphSAGE learns inductive neighborhood aggregators that generalize to unseen nodes~\cite{hamilton2017graphsage}. 

\paragraph{Implication for road networks.} Human movement on streets is directional and constrained by intersection geometry; we therefore combine strong structural priors (Node2Vec) with an orientation‑aware, sectorized pooling that injects distance/bearing/density semantics into descriptors, complementing GraphSAGE‑style local aggregation.

\section*{Methodology}

\begin{figure}[!t]
  \centering
  \includegraphics[width=\columnwidth]{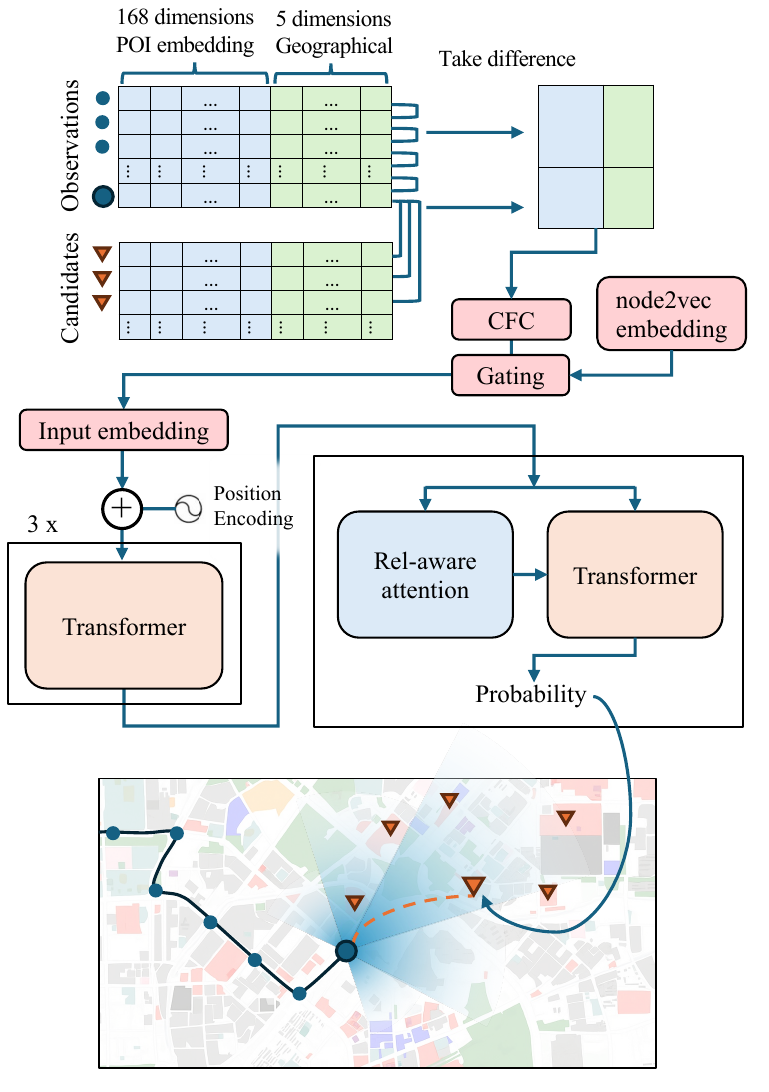}
  \caption{\textbf{Overview of our Relation‑Aware LNN–Transformer.}
  A road‑user trajectory is projected onto the intersection graph and encoded via sector‑wise POI and geometric features fused with Node2Vec embeddings.  
  The differenced features go through a CfC‑LNN and $L\!-\!1$ standard Transformer layers, followed by a relation‑aware attention block that biases self‑attention with inter‑node bearings, and finally a softmax over one‑hop candidates.}
  \label{fig:overview}
\end{figure}

As Figure~\ref{fig:overview} illustrates our pipeline, every trajectory is first snapped to intersections and enriched with POI and geometric features. Then, these features are differenced and processed by our Relation-Aware LNN–Transformer architecture. Finally, a softmax head over one‑hop neighbors produces the next‑node prediction.  

\subsection{Problem Definition}

Let \(G=(V,E)\) be an undirected road graph whose nodes \(V\) are intersections and edges \(E\) are road segments. 
Each node \(v\) has a structural embedding \(\mathbf{h}_v^{\mathrm{struct}}\in\mathbb{R}^{d_s}\) and a POI descriptor \(\mathbf{h}_v^{\mathrm{POI}}\in\mathbb{R}^{d_p}\). 
Let \(\mathbf{p}_v\in\mathbb{R}^2\) denote its planar coordinates (fixed coordinate reference system).

Given a trajectory \(\tau=(v_1,\dots,v_T)\), we predict the next node \(v_{T+1}\) from the one‑hop candidate set \(\mathcal{C}(v_T)=\mathcal{N}(v_T)\). 
We pass \(\tau\) through our encoder (Secs.~\ref{sec:poi}-\ref{sec:rat}) to obtain a representation \(\mathbf{z}_u\) for each \(u\in\mathcal{C}(v_T)\), and score
\begin{equation}
\begin{split}
s_\theta(u \mid \tau)
&= \mathbf{w}^\top \mathbf{z}_u + b,\\
p_\theta(u \mid \tau)
&= \operatorname{softmax}\bigl(s_\theta(u \mid \tau)\bigr)
   \quad\text{for }u\in\mathcal{N}(v_T)\,.
\end{split}
\end{equation}
We define $\mathcal{L}_{\mathrm{CE}}$, the classification loss as follows:
\begin{equation}
\mathcal{L}_{\mathrm{CE}}
= -\tfrac{1}{N}\!\sum_{i=1}^N\log p_\theta\!\bigl(v_{T+1}^{(i)}\mid\tau^{(i)}\bigr).
\end{equation}
We also regress the unit direction vector from \(v_T\) to the ground‑truth next node,
\(\mathbf{d}_{T\to T+1}=\frac{\mathbf{p}_{T+1}-\mathbf{p}_{T}}{\lVert \mathbf{p}_{T+1}-\mathbf{p}_{T}\rVert}\),
using a cosine loss
\begin{equation}
\mathcal{L}_{\mathrm{dir}}
= \tfrac{1}{N}\!\sum_{i=1}^N \bigl[1-\hat{\mathbf{d}}_{T\to T+1}^{(i)}\!\cdot\!\mathbf{d}_{T\to T+1}^{(i)}\bigr].
\end{equation}
The overall objective is \(\mathcal{L}=\mathcal{L}_{\mathrm{CE}}+\gamma\,\mathcal{L}_{\mathrm{dir}}\), with \(\gamma>0\).

\subsection{Sector-Wise Directional POI Aggregation}
\label{sec:poi}
Let \(\mathcal{C}\) be the set of POI categories (\(|\mathcal{C}|=12\)). 
Around each node \(v\), we consider a circular neighborhood of radius \(R=150\,\mathrm{m}\)\footnote{Radius selection was based on ablation over \(R\in\{50,75,100,125,150,175,200,225,250\}\) m: \(R=150\) m achieves a favorable trade-off—coverage \(\approx0.63\), duplication \(\approx0.52\), and marginal gain \(\approx0.62\)—and yields the highest Test@1 (0.9094) and MRR (0.9506) in our experiments.} and partition it into \(S=8\) equal angular sectors.%
\footnote{Varying sectors \(S\in\{2,4,8,16\}\) yields optimal performance at \(S=8\): Test@1=0.9094, MRR=0.9506 (vs.\ 0.8734/0.9237 at \(S=2\) and 0.8704/0.9232 at \(S=16\)).}

For each category \(c\in\mathcal{C}\), we compute a 14-dimensional summary \(f_v^{(c)}\) consisting of:

\paragraph{(i) Category-wise circular statistics (5).}
Given POIs \(\{(d_i,\theta_i)\}_i\) within \(R\), we compute
\begin{equation}
\begin{aligned}
\mu_d      &= \frac{1}{n}\sum_i d_i, 
&\qquad
\sigma_d^2 &= \frac{1}{n}\sum_i (d_i-\mu_d)^2,\\
m_c        &= \frac{1}{n}\sum_i \cos\theta_i, 
&\qquad
m_s        &= \frac{1}{n}\sum_i \sin\theta_i,\\
r          &= \sqrt{m_c^2 + m_s^2}\,.
\end{aligned}
\end{equation}

\paragraph{(ii) Angular sector densities (8).}
For sector \(k\in\{1,\dots,S\}\), 
\(h_k=\tfrac{1}{A_S}\,\bigl|\{\theta_i\in\mathrm{bin}_k\}\bigr|\), 
where \(A_S=\pi R^2/S\).

\paragraph{(iii) Presence indicator (1).}
A binary flag for “existence of at least one POI of category \(c\) within \(R\)”.

\noindent
Concatenating over categories gives \(\mathbf{x}_v=[\,f_v^{(c)}\,]_{c\in\mathcal{C}}\in\mathbb{R}^{168}\).
We include a per‑feature binary mask for empty categories/sectors and apply z‑score normalization fitted on the training set; features are precomputed offline and cached.

\subsection{CfC–LNN Temporal Encoder}
\label{sec:cfc}
To expose instantaneous dynamics, we feed first‑order differences of POI and geometric features into a Continuous‑time Forgetting Cell (CfC): 
\(\Delta\mathbf{p}_t=\mathbf{p}_t-\mathbf{p}_{t-1}\), 
\(\Delta\mathbf{g}_t=\mathbf{g}_t-\mathbf{g}_{t-1}\) (zeros at \(t{=}1\)). 
Let \(\mathbf{h}_t\in\mathbb{R}^{d_h}\) be the latent state. 
We use a closed‑form decay with a \emph{constant} time step (our data do not require explicit \(\Delta t\) modeling), 
\begin{equation}
\begin{aligned}
\alpha 
&= \exp\!\bigl(-\operatorname{softplus}(\boldsymbol{\tau})\bigr), \\[6pt]
[\mathbf{a}_t;\mathbf{b}_t] 
&= W\bigl[\Delta\mathbf{p}_t;\,\Delta\mathbf{g}_t;\,\mathbf{h}_{t-1}\bigr] + \mathbf{b}, \\[6pt]
\mathbf{c}_t 
&= \tanh(\mathbf{b}_t), \\[6pt]
\mathbf{h}_t 
&= \alpha \odot \mathbf{h}_{t-1}
   \;+\;(1 - \alpha)\odot \mathbf{c}_t.
\end{aligned}
\end{equation}
We then combine \(\mathbf{h}_t\) with the static structural prior via a gated mixer
\begin{equation}
\begin{aligned}
\mathbf{g}_t 
&= \sigma\!\bigl(W_g[\mathbf{h}_t \Vert \mathbf{h}^{\mathrm{struct}}_{v_t}] + \mathbf{b}_g\bigr), \\[4pt]
\tilde{\mathbf{h}}_t 
&= \mathbf{g}_t \odot \mathbf{h}^{\mathrm{struct}}_{v_t}
   + (1 - \mathbf{g}_t)\odot \mathbf{h}_t.
\end{aligned}
\end{equation}

and project to the model dimension \(d\).
After encoding the observed steps \(1{:}T\), we append one candidate‑specific state per \(u_j\in\mathcal{N}(v_T)\) by a final CfC update initialized at \(\mathbf{h}_T\) and conditioned on the geometry from \(v_T\) to \(u_j\), yielding
\begin{equation}
    \mathbf{Z}=\bigl[\tilde{\mathbf{h}}_1,\dots,\tilde{\mathbf{h}}_T,\,\tilde{\mathbf{h}}_{T+1}^{(1)},\dots,\tilde{\mathbf{h}}_{T+1}^{(C)}\bigr]\in\mathbb{R}^{(T+C)\times d}.
\end{equation}

Since LNNs favor input‑modulated continuous dynamics, turning static POI context into incremental signals matches this inductive bias. 
Empirically, removing differencing reduces performance from \(\text{Test@1}=0.9094,\,\text{MRR}=0.9506\) to \(0.8752,\,0.9259\) (8 sectors; 3 standard layers + 1 relation‑aware layer).

\subsection{Relation‑Aware Transformer}
\label{sec:rat}
We process \(\mathbf{Z}\) with a Transformer encoder augmented by directional and type biases.

\paragraph{Bearing‑aware additive bias.}
For tokens \(p,q\) corresponding to nodes \(i_p,i_q\), let \(\theta_{p\to q}\) be the bearing from \(\mathbf{p}_{i_p}\) to \(\mathbf{p}_{i_q}\).
For each head \(h\), we form a head‑specific additive bias
\begin{equation}
B^{(h)}_{pq}=\lambda_h\,[\cos\theta_{p\to q},\,\sin\theta_{p\to q}]\,\mathbf{w}_h,
\end{equation}
and compute
\begin{equation}
\operatorname{Attn}^{(h)}(Q,K,V)=\operatorname{softmax}\!\Bigl(\tfrac{QK^\top}{\sqrt{d_h}}+B^{(h)}\Bigr)V,
\end{equation}
where \(d_h=d/H\), \(\lambda_h\) is learned and row‑centered for numerical stability. 
This captures directional alignment while letting heads specialize.

\paragraph{Type embeddings and masking.}
We add learned type embeddings to distinguish \emph{observed} vs.\ \emph{candidate} tokens and apply an attention mask that blocks candidate\(\leftrightarrow\)candidate attention (candidates attend to history only). 
All other components (residual, pre‑norm, feed‑forward) follow the standard encoder.

\section*{Experiment}

\subsection{Dataset \& Pre‐processing}
Our experiments use two sources: (i) a static road--POI graph extracted from OpenStreetMap (OSM)\footnote{\url{https://www.openstreetmap.org}}
%\cite{openstreetmap2025}
and (ii) road‐user GPS trajectories from GeoLife~\cite{zheng2009mining,zheng2008understanding,zheng2010geolife}. 
GeoLife trajectories are collected by phones and dedicated GPS loggers with diverse sampling rates; notably, \(\sim\)91.5\% are densely sampled (every \(1\!-\!5\)s or \(5\!-\!10\)m per point), while the constructed intersection graph is relatively sparse (mean inter‐intersection spacing \(\approx\)487\,m). 
This density/sparsity contrast motivates a topology‐respecting projection that \emph{does not} impute unobserved paths with global shortest routes.

\paragraph{Static Road--POI Graph.}
We obtain the urban road network with Overpass Turbo~\cite{overpass-turbo}, treating intersections as nodes \(V\) and road segments as edges \(E\).
Each POI is assigned to its nearest node, and we apply our Sector‐Wise Directional POI Aggregation (eight angular sectors) to form a compact 168‐dimensional descriptor per node (Sec.~\ref{sec:poi}); features are z‐score normalized on the training set and cached.
This yields two per‐node vectors: \(\mathbf{h}_v^{\mathrm{struct}}\in\mathbb{R}^{d_s}\) and \(\mathbf{h}_v^{\mathrm{POI}}\in\mathbb{R}^{d_p}\).

\paragraph{GPS‐to‐Graph Projection (Topology‐respecting, no path imputation).}
GeoLife provides time‐stamped \((\text{lat},\text{lon},t)\) samples for road users (walking, cycling, driving). 
We convert raw GPS streams to node sequences on \(G\) via four steps that preserve observed geometry and avoid shortest‐path completion:
\begin{enumerate}
  \item \textbf{Edge snapping.} 
  Each GPS point is snapped to its nearest road \emph{edge} (within a search radius), yielding a continuous position along a specific segment; ties are broken by heading consistency.
  \item \textbf{Intersection hit detection.}
  We detect an intersection visit when the snapped position enters a fixed buffer around node \(v\) (or crosses its Voronoi cell boundary). 
  A small hysteresis window prevents rapid oscillation between adjacent nodes.
  \item \textbf{Twig pruning (jitter removal).}
  We remove short detours caused by GPS jitter: any subsequence that leaves a main path and returns to the same node within two hops is pruned.
  \item \textbf{Repeat preservation (no de‐dup).}
  We \emph{do not} collapse consecutive occurrences of the same node. 
  Repeated visits (low speed or dwell near an intersection) are retained in the observed context, providing implicit temporal/dwell cues even without explicit time encoding. 
  Note that labels \(v_{T+1}\) are always different from \(v_T\) by construction (prediction is over one‐hop neighbors).
\end{enumerate}

\paragraph{Segmentation and Candidate Construction.}
We segment each snapped stream into maximal contiguous sequences that remain within the mapped city graph and pass basic quality filters (e.g., plausible speed bounds; removal of isolated outliers).
For each step \(T\), the candidate set is the one‐hop neighborhood \(\mathcal{C}(v_T)=\mathcal{N}(v_T)\); the ground‐truth next node \(v_{T+1}\) is guaranteed to lie in \(\mathcal{C}(v_T)\).
We compute evaluation metrics on this neighbor‐ranking task.

\paragraph{Trajectory Segmentation}  
We partition each projected trajectory into non-overlapping segments by sliding a fixed node-count window chosen to approximate four spatial scales: short-range ($\approx$1\,km), mid-range ($\approx$3\,km), long-range ($\approx$5\,km), and extended ($\ge$7\,km). Table~\ref{tab:traj-segments} summarizes the number of segments at each scale, their exact mean lengths, and the corresponding node-count ranges.

\begin{table}[h]
  \centering
  \small
  \resizebox{\columnwidth}{!}{%
  \begin{tabular}{lllll}
    \toprule
    \textbf{Range Category}   & \textbf{Segments} & \textbf{Mean Length (m)} & \textbf{Node-Count Range} \\ 
    \midrule
    Short-range ($\approx$1 km)       & 63,813            & 1,028                    & 7–20                      \\
    Mid-range ($\approx$3 km)         & 29,941            & 2,983                    & 20–40                     \\
    Long-range ($\approx$5 km)        & 16,662            & 5,047                    & 40–100                    \\
    Extended ($\geq$7 km)     & 3,453             & 7,215                    & 100–256                   \\ 
    \midrule
    Full trajectories         & 16,624            & 9,842                    & 2–769                     \\ 
    \bottomrule
  \end{tabular}
  }
  \caption{Counts and statistics for trajectory segments by spatial range category.}
  \label{tab:traj-segments}
\end{table}

\subsection{Evaluation Metrics}

In our setting, the next node is selected from a discrete candidate set \(\mathcal{C}(v_T)\). Practical systems act on only the top few options, so we emphasize \emph{actionable precision} via Acc@\(k\). When top-1 is wrong, we still require the correct node to receive higher relative confidence (i.e., appear near the top), which MRR captures. Finally, because class balance varies across neighborhoods and thresholds are arbitrary, we report mean ROC–AUC as a threshold-free measure of separability in one-vs-rest scoring.

\begin{enumerate}
  \item \textbf{Accuracy@k} (\(k=1,3,5\)).  
  Given scores \(\{s_u\}_{u\in\mathcal{C}(v_T)}\), let \(\mathrm{rank}(v_{T+1})\) be the position of the correct node in descending order. Then
  \[
    \mathrm{Acc}@k
    = \frac{1}{|\mathcal{D}|}\sum_{(v_1,\dots,v_{T+1})\in\mathcal{D}}
      \mathbf{1}\bigl(\mathrm{rank}(v_{T+1})\le k\bigr).
  \]
  \item \textbf{Mean Reciprocal Rank (MRR).}  
  \[
    \mathrm{MRR}
    = \frac{1}{|\mathcal{D}|}\sum_{(v_1,\dots,v_{T+1})\in\mathcal{D}}
      \frac{1}{\mathrm{rank}(v_{T+1})}.
  \]
  \item \textbf{Mean ROC–AUC.}  
  Average area under the ROC curve across one-vs-rest tasks (each candidate vs.\ all others), giving threshold-free discrimination under class imbalance.
\end{enumerate}

\begin{table*}[t!]
  \centering
  \small
  
  \resizebox{\textwidth}{!}{%
  \begin{tabular}{lccccclccc}
    \toprule
    \textbf{Model} & \textbf{Layers} & \textbf{Heads} & \textbf{Hidden/Dim} & \textbf{FFN/GCN} & \textbf{Specifc Features} & \textbf{POI} & \textbf{Node2Vec} & \textbf{Geo} & \textbf{Params (M)} \\
    \midrule
    LSTM (1997)            & 2                 & –   & 512              & –                & Bahdanau attention over final states          & \cmark & \cmark & \cmark & 2.46 \\
    HST\mbox{-}LSTM (2018) & GRU+LSTM          & –   & 224/224          & –                & neighbor aggregation \(k{=}8\)                & \cmark & \cmark & \xmark & 2.43 \\
    ARNN (2020)            & 1 (GRU)           & –   & 384              & –                & attention over \(k{=}8\) Node2Vec neighbors   & \cmark & \cmark & \xmark & 2.15 \\
    MobGT (2023)           & 3                 & 4   & 192              & 192              & graph Transformer                             & \cmark & \cmark & \cmark & 2.45 \\
    STGN (2019)            & 2 blocks          & –   & 512 (GRU)        & 384\(\to\)512    & spatio-temporal GRU+graph conv (edge attrs)   & \cmark & \cmark & \cmark & 2.59 \\
    GETNext (2023)         & 6                 & 4   & 256              & 192              & Transformer + trajectory flow map             & \cmark & \cmark & \cmark & 2.58 \\ \midrule
    \textbf{Our Model (LNN+RelT)}   & 4                 & 4   & 256              & 256              & CfC–LNN encoder + bearing-biased attention; \(\gamma{=}0.1\) & \cmark & \cmark & \cmark & 2.34 \\
    \bottomrule
  \end{tabular}
  }
    \caption{Baseline configurations. “–” denotes not applicable. All methods share the same training protocol; inputs indicate whether POI histograms, Node2Vec, and geometric features are used.}
    \label{tab:baseline-config}
\end{table*}

\paragraph{Baseline Models}
We compare against six classical and recent methods that, like ours, operate on discrete node sequences over an abstract graph (LSTM) to cover different paradigms of graph-based trajectory modeling: LSTM (vanilla recurrent sequence encoder)~\cite{hochreiter1997lstm}; HST-LSTM (hierarchical spatio-temporal LSTM)~\cite{kong2018hstlstm}; STGN (spatio-temporal gated network)~\cite{zhao2019stgn}; ARNN (attentional RNN)~\cite{guo2020arnn}; MobGT ~\cite{xu2023mobgt}; and GETNext (transformer augmented with a trajectory-flow map)~\cite{yang2023getnext}.
All models use the same 8:1:1 train/val/test split and training protocol (batch size 16, 10 epochs, learning rate \(1\!\times\!10^{-4}\)).  
Table~\ref{tab:baseline-config} summarizes model configurations and inputs.  

\begin{table*}[t]
\centering

\small
\resizebox{\textwidth}{!}{%
\renewcommand{\arraystretch}{1.30}
\begin{tabular}{l c
  *{4}{c@{\hspace{4pt}}}|
  *{4}{c@{\hspace{4pt}}}|
  *{4}{c@{\hspace{4pt}}}|
  *{4}{c@{\hspace{4pt}}}|
  *{4}{c}
}
\toprule
\multirow{2}{*}{\bfseries Model} & \multirow{2}{*}{\bfseries Mean AUC}
 & \multicolumn{4}{c|}{\bfseries 1\,km}
 & \multicolumn{4}{c|}{\bfseries 3\,km}
 & \multicolumn{4}{c|}{\bfseries 5\,km}
 & \multicolumn{4}{c|}{\bfseries 7\,km}
 & \multicolumn{4}{c}{\bfseries Full (10\,km)} \\
\cmidrule(lr){3-6}\cmidrule(lr){7-10}\cmidrule(lr){11-14}\cmidrule(lr){15-18}\cmidrule(lr){19-22}
 &  & Acc@1 & Acc@3 & Acc@5 & MRR
        & Acc@1 & Acc@3 & Acc@5 & MRR
        & Acc@1 & Acc@3 & Acc@5 & MRR
        & Acc@1 & Acc@3 & Acc@5 & MRR
        & Acc@1 & Acc@3 & Acc@5 & MRR \\
\midrule
LSTM (1997)            & 0.9194
 & 0.7735 & 0.9279 & 0.9817 & 0.8571
 & 0.6631 & 0.8735 & 0.9643 & 0.7815
 & 0.7642 & 0.9118 & 0.9748 & 0.8465
 & 0.7197 & 0.8671 & 0.9566 & 0.8107
 & 0.7318 & 0.9050 & 0.9741 & 0.8274 \\
STGN (2019)            & 0.8837
 & 0.5839 & 0.8655 & 0.9624 & 0.7400
 & 0.5586 & 0.8244 & 0.9533 & 0.7140
 & 0.6071 & 0.8596 & 0.9586 & 0.7585
 & 0.5867 & 0.8006 & 0.9364 & 0.7129
 & 0.6037 & 0.8521 & 0.9543 & 0.7530 \\
MobGT (2023)           & 0.7962
 & 0.4615 & 0.8343 & 0.9581 & 0.6624
 & 0.4417 & 0.7943 & 0.9249 & 0.6344
 & 0.4457 & 0.7439 & 0.9118 & 0.6242
 & 0.4277 & 0.7312 & 0.8584 & 0.6068
 & 0.3277 & 0.6777 & 0.8978 & 0.5438 \\
HST\mbox{-}LSTM (2018) & 0.8876
 & 0.6221 & 0.8655 & 0.9631 & 0.7559
 & 0.5342 & 0.7810 & 0.9446 & 0.6840
 & 0.6653 & 0.8794 & 0.9604 & 0.7832
 & 0.6936 & 0.8382 & 0.9451 & 0.7890
 & 0.6500 & 0.8683 & 0.9549 & 0.7713 \\
ARNN (2020)            & 0.9199
 & \underline{0.8058} & 0.9567 & 0.9901 & \underline{0.8838}
 & 0.6674 & 0.8775 & 0.9706 & 0.7857
 & 0.7516 & 0.9202 & 0.9778 & 0.8426
 & 0.7341 & 0.8844 & 0.9653 & 0.8238
 & 0.7288 & 0.9050 & 0.9765 & 0.8259 \\
GETNext (2023)         & \textbf{0.9471}
 & 0.7849 & \underline{0.9599} & \underline{0.9932} & 0.8730
 & \underline{0.7706} & \underline{0.9402} & \underline{0.9886} & \underline{0.8600}
 & \underline{0.8266} & \underline{0.9448} & \underline{0.9874} & \underline{0.8908}
 & \underline{0.7775} & \underline{0.9017} & \underline{0.9740} & \underline{0.8548}
 & \underline{0.7919} & \underline{0.9327} & \underline{0.9844} & \underline{0.8687} \\ \midrule
\bfseries Our Model (LNN+RelT) & \bfseries 0.9302
 & \bfseries 0.9104 & \bfseries 1.0000 & \bfseries 1.0000 & \bfseries 0.9517
 & \bfseries 0.9008 & \bfseries 0.9952 & \bfseries 1.0000 & \bfseries 0.9443
 & \bfseries 0.9011 & \bfseries 1.0000 & \bfseries 1.0000 & \bfseries 0.9511
 & \bfseries 0.8844 & \bfseries 0.9877 & \bfseries 1.0000 & \bfseries 0.9350
 & \bfseries 0.9094 & \bfseries 1.0000 & \bfseries 1.0000 & \bfseries 0.9506 \\
\midrule
\textbf{Rel. Improv. (\%)} 
 & 
 & 12.98 & 4.18 & 0.69 & 7.68
 & 16.90 & 5.85 & 1.15 & 9.80
 &  9.01 & 5.84 & 1.28 & 6.77
 & 13.75 & 9.54 & 2.67 & 9.39
 & 14.83 & 7.21 & 1.59 & 9.42 \\
\bottomrule
\end{tabular}%
}
\caption{Performance comparison of next–node prediction across varying average path lengths. We report Acc@1/3/5 and MRR. The last row shows the relative improvement of our model noted LNN+RelT over the best baseline (\%).}
\label{tab:path_length_performance}
\end{table*}

\subsection{Comparisons with Models}

Table~\ref{tab:path_length_performance} reports our model’s performance against six baselines across different path lengths and we can get the four following conclusions. 

\paragraph{Fair Capacity and Global Gains.}
With only 2.34M parameters, our model consistently surpasses the best baseline across all lengths and metrics.
All Acc@1 improvements are statistically significant at the 95\% confidence level ($p<0.05$; paired bootstrap over test sessions).
Relative Acc@1 gains range from 9.0\% (5\,km) to 17.0\% (3\,km), averaging 12.2\% across cohorts.

\paragraph{AUC under Class Imbalance.}
While our top-$k$ ranking is strongest overall, the mean ROC--AUC column shows that GETNext achieves the highest threshold-free discrimination (0.9471) and our model ranks a close second (0.9302).
This gap indicates that, under severe class imbalance and across varying thresholds, GETNext maintains a slightly better global TPR--FPR trade-off.
Nevertheless, our method converts probability mass more decisively to the top of the candidate list (higher Acc@1/3/5 and MRR), which is crucial for next-node selection where only the top few suggestions are actionable.

\paragraph{Robust Mid-Range Performance.}
The largest relative uplifts occur at 3\,km (Acc@1 $\uparrow$ 17.0\%, MRR $\uparrow$ 9.9\%) and 7\,km (Acc@1 $\uparrow$ 13.7\%, Acc@3 $\uparrow$ 9.5\%), highlighting the benefit of our Relation-Aware LNN--Transformer: CfC captures local, rate-like changes, while bearing-biased attention aligns long-range directional dependencies on the graph.

\paragraph{Unmatched Top-$k$ Recall.}
Beyond 1\,km, our model retains perfect Acc@5 ($1.000$) for all lengths, whereas the best baselines dip below $0.99$.
For navigation or dispatch-style applications, such zero miss rate in the top shortlist translates to more reliable candidate generation before downstream reranking or planning.

\subsection{Transformer Depth \& Head Sensitivity}

\paragraph{Search Grid Design.}
We sweep Transformer depth \(L\in\{1,2,4,6,8\}\) and heads \(H\in\{2,4,8,16\}\) on the validation split for both our model and the strongest baseline, \textbf{GETNext}.  
These values follow a power‐of‐two progression (with an intermediate depth of 6) to cover under‑parameterized, mid‑range, and over‑parameterized regimes in a tractable grid, allowing us to assess minimal viable complexity, saturation behavior, and computational trade‑offs.  

\begin{figure*}[t]
  \centering
  \begin{subfigure}[b]{0.48\textwidth}
    \centering
    \includegraphics[width=.8\textwidth]{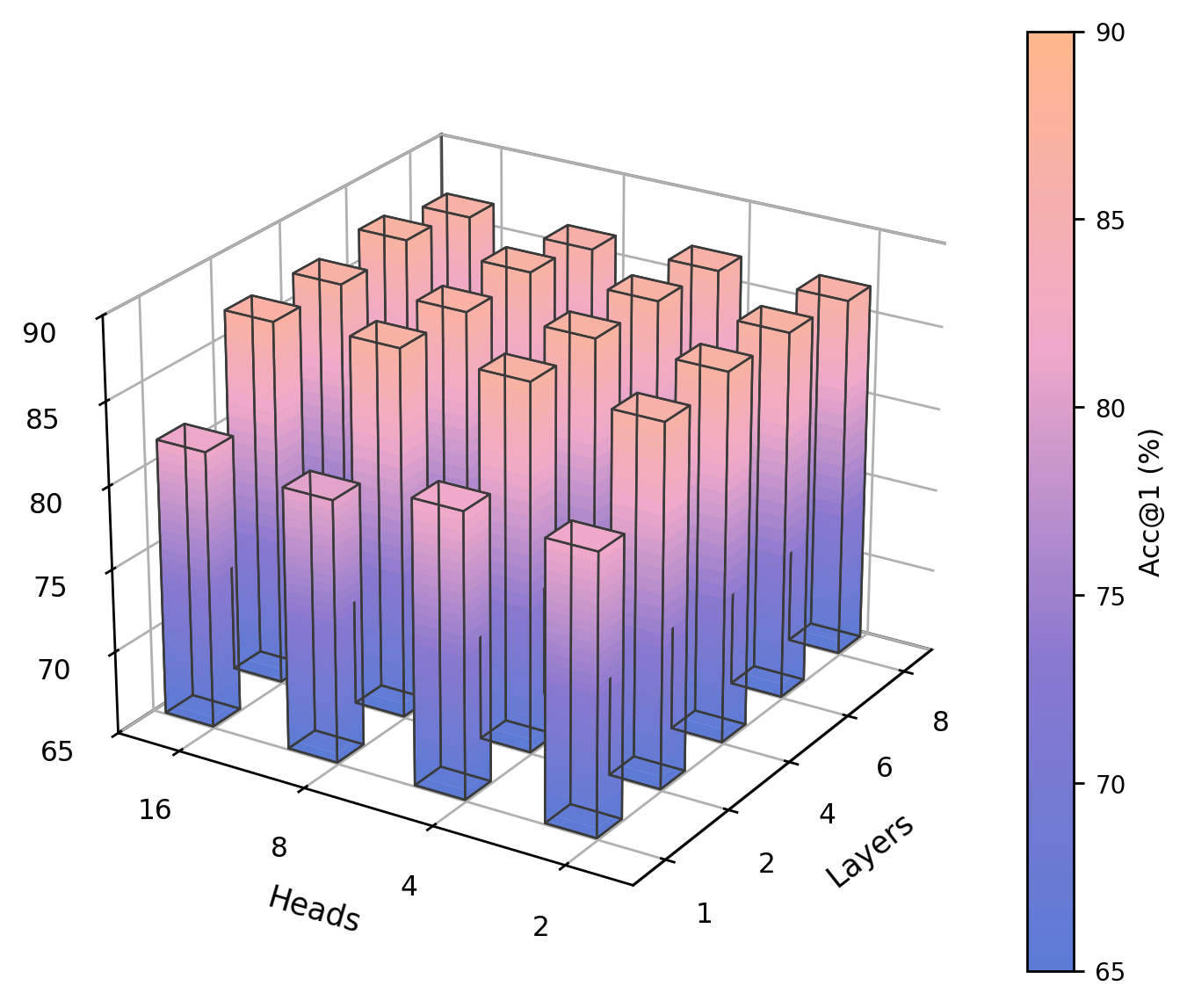}
    \caption{LNN+RelT Sensitivity}
    \label{fig:hp-sensitivity-my}
  \end{subfigure}
  \hfill
  \begin{subfigure}[b]{0.48\textwidth}
    \centering
    \includegraphics[width=.8\textwidth]{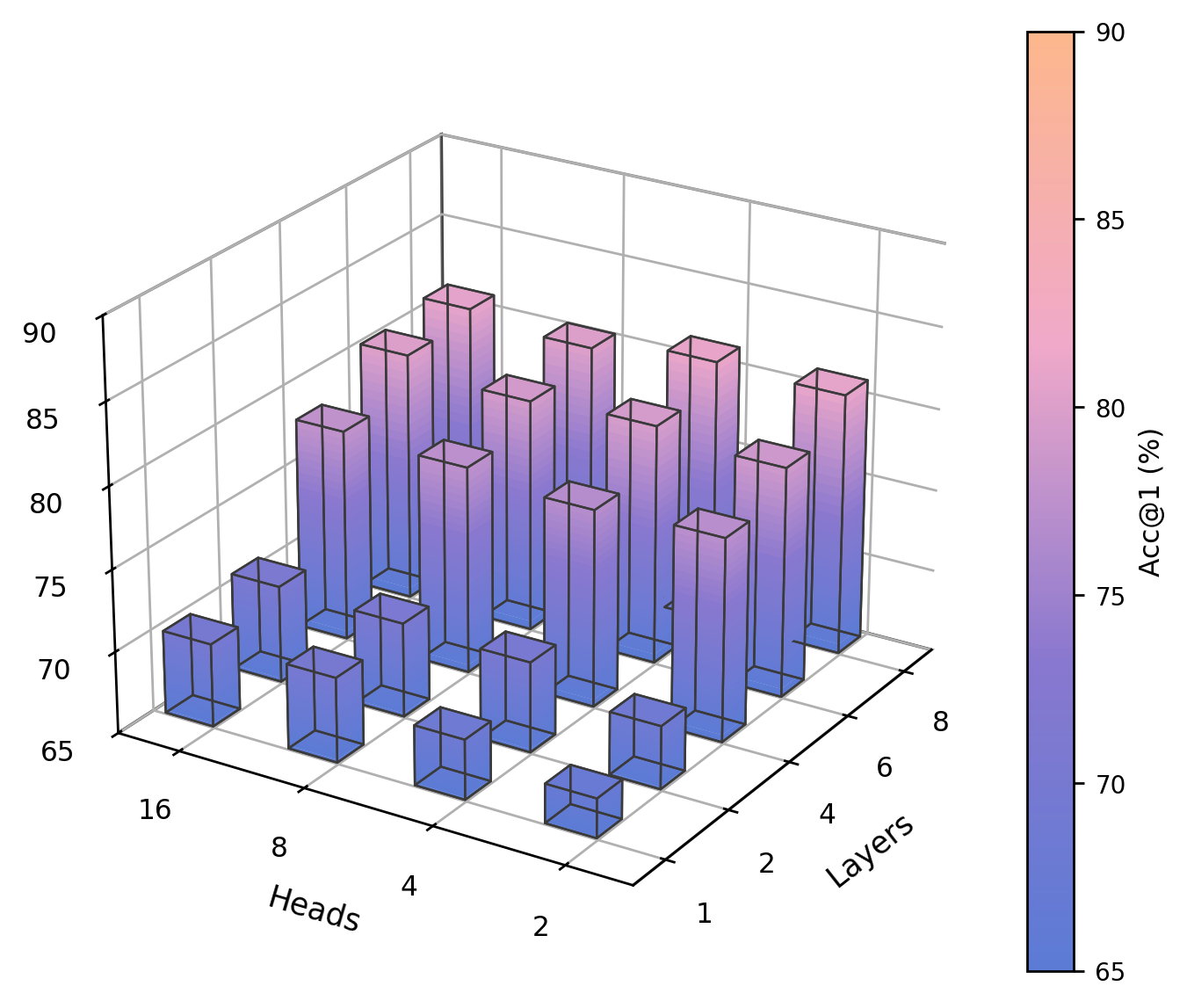}
    \caption{GETNext Sensitivity}
    \label{fig:hp-sensitivity-getnext}
  \end{subfigure}
  \caption{Validation Acc@1 as a function of Transformer layers ($L$) and heads ($H$).}
  \label{fig:hp-sensitivity}
\end{figure*}

We select GETNext as our comparison baseline for this part because (i) it achieves the highest mean ROC–AUC and competitive Acc/MRR in Table~\ref{tab:path_length_performance}, and (ii) like our model, it is Transformer-based with independently tunable depth ($L$) and head count ($H$), enabling a directly comparable sensitivity study.
Figure~\ref{fig:hp-sensitivity} visualizes Acc@1 across the grid.

\paragraph{Key observations.}
\begin{itemize}
    \item \textbf{Early saturation. }
  Our LNN+RelT model surpasses \textbf{87\%} Acc@1 with only $L{=}2$, $H{=}4$ (\(0.8714\)), and is already near-peak at $L{=}4$, $H{=}4$ (\(0.8731\)); the best point in the grid is \(0.8741\) at $L{=}6$, $H{=}16$.
  Beyond $L{\ge}2$, all changes are within \(\le\)0.5 percentage points, indicating low sensitivity to over-parameterization.

  \item \textbf{Monotonic gains.}
  GETNext improves steadily with depth: from \(0.7050\) at $L{=}2$, $H{=}4$ to \(0.8122\) at $L{=}8$, $H{=}4$, with the largest gains between $L{=}2{\to}6$ (e.g., \(0.8169{\to}0.8798\) at $H{=}4$). 
  This suggests GETNext relies more on deeper self-attention stacks to realize long-range dependencies learned from trajectory flows.

  \item \textbf{Accuracy–efficiency trade-off.}
  Achieving \(\approx\)0.873 Acc@1 at $L{=}4$, $H{=}4$ (LNN+RelT) versus GETNext requiring $L{=}8$ to exceed 0.81 implies fewer attention blocks and heads for comparable (or better) accuracy.
  Since attention cost scales roughly with $L\!\times\!H$, our near-peak setting halves compute relative to deeper GETNext settings, supporting faster inference without sacrificing top-$k$ ranking quality.
\end{itemize}

\subsection{Perturbation Sensitivity}
\begin{figure}[t!]
  \centering
  \includegraphics[width=\columnwidth]{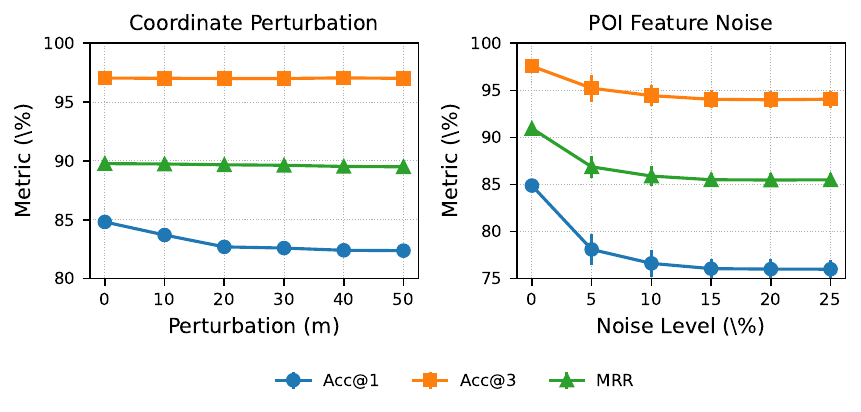}
  \caption{Robustness under coordinate perturbation and POI-feature noise.}
  \label{fig:robustness}
\end{figure}
We evaluate robustness under two perturbations and visualize the outcomes in Fig.~\ref{fig:robustness}.
For each noise level we run $10$ trials and report mean$\pm$std for Acc@1/3/5 and MRR.

\paragraph{Protocol.}
(i) \emph{Coordinate noise:} Add i.i.d.\ zero-mean Gaussian displacements to each GPS sample before projection, with standard deviation $\sigma\in\{0,10,20,30,40,50\}$\,m; paths are then re-projected.
(ii) \emph{POI-feature noise:} Apply multiplicative Gaussian perturbations independently to each of the 8-sector POI bin counts:~$\tilde{h}_k = h_k\,(1+\varepsilon_k)$ with $\varepsilon_k\sim\mathcal{N}(0,\sigma^2)$ and $\sigma\in\{0,0.05,0.10,0.15,0.20,0.25\}$;
we clip negatives to zero and renormalize within a node to preserve total mass.

\paragraph{Perturbation Sensitivity Results}
\begin{itemize}
  \item \textbf{Spatial stability.}
  Acc@1 decreases from $0.8481\!\pm\!0.0013$ ($\sigma{=}0$) to $0.8237\!\pm\!0.0038$ ($\sigma{=}50$\,m), a total drop of $2.44$\,pp ($\approx0.49$\,pp per $10$\,m).
  Acc@3 remains essentially flat ($0.9704\!\to\!0.9702$; $\leq0.05$\,pp per $10$\,m), and MRR changes by only $0.27$\,pp ($0.8977\!\to\!0.8950$).
  Acc@5 stays at $1.0000$ across all $\sigma$.
  These trends indicate that graph-structural cues and bearing-aware attention buffer realistic coordinate jitter.

  \item \textbf{Semantic sensitivity (early-drop, then plateau).}
  Under POI-feature noise, the largest degradation occurs at the first step:
  Acc@1 from $0.8488\!\pm\!0.0009$ ($\sigma{=}0$) to $0.7809\!\pm\!0.0162$ ($\sigma{=}0.05$), i.e., $-6.79$\,pp; Acc@3 from $0.9757$ to $0.9523$ ($-2.34$\,pp); MRR from $0.9095$ to $0.8687$ ($-4.08$\,pp).
  Beyond $\sigma{=}0.10$, the curves flatten (e.g., Acc@1 $0.7661{\to}0.7599$ from $10\%$ to $25\%$), suggesting the model retains a stable ranking signal once sector histograms are moderately corrupted.

  \item \textbf{Complementary resilience.}
  Contrasting the two stressors shows that \emph{spatial} perturbations barely affect top-$k$ recall, whereas \emph{semantic} corruption primarily impacts Acc@1/MRR.
  This aligns with our architecture: Node2Vec structural priors and road-graph bearings anchor the candidate set, while sector-wise POI descriptors act as high-gain semantic cues.
\end{itemize}

\noindent
Overall, the integration of structural embeddings, directional POI aggregation, and CfC dynamics ensure model's high tolerance to positional jitter and semantic noise—desirable properties for noisy, real-world navigation deployments.

\subsection{Ablation Study}

We conduct four ablations to quantify each component’s impact: (1) feature-branch removal, (2) layer composition,  (3) POI aggregation granularity and (4) Radius Analysis

\subsubsection{Feature-Branch Removal}
As shown in Table~\ref{tab:ablation-features}, removing all POI inputs causes the largest drop (Acc@1 –32.8 pp, MRR –19.3 pp), indicating that raw POI counts are the primary semantic signal. Without POI-diff (\cmark\textsuperscript{*}), the model loses a further 3.4 pp in Acc@1, showing fine-grained \(\Delta\)-features provide a modest but consistent boost.  Omitting geometric deltas shrinks Acc@1 by 8.9 pp (MRR –6.2 pp), and dropping structural (Node2Vec) embeddings costs 19.3 pp in Acc@1, confirming that spatial context and global topology are both critical to robust next-node prediction.  
Altogether, each branch contributes meaningfully, with POI counts as the most essential component.  
\begin{table}[t]
  \centering
  \small

  \resizebox{\columnwidth}{!}{%
    \begin{tabular}{ccc|cccc}
      \toprule
      POI & Geo & Node2Vec
        & Acc@1 & Acc@3 & Acc@5 & MRR \\
      \midrule
      \cmark & \cmark & \cmark & 0.9094 & 1.0000 & 1.0000 & 0.9506 \\
      \xmark & \cmark & \cmark & 0.5819 & 1.0000 & 1.0000 & 0.7573 \\
      \cmark & \xmark & \cmark & 0.8200 & 0.9328 & 0.9820 & 0.8844 \\
      \cmark & \cmark & \xmark & 0.7121 & 1.0000 & 1.0000 & 0.8354 \\
      \cmark* & \cmark & \cmark & 0.8752 & 0.9802 & 1.0000 & 0.9259 \\
      \bottomrule
    \end{tabular}
  }
  \caption{Feature-branch ablation on the full test set.
  \label{tab:ablation-features}
  (\cmark\textsuperscript{*} indicates removal of differential POI features).}
\end{table}

\begin{table}[H]
  \centering
  \small

  \resizebox{\columnwidth}{!}{%
    \begin{tabular}{cc|cccc}
      \toprule
      Std & Rel-Aware & Acc@1 & Acc@3 & Acc@5 & MRR \\
      \midrule
      4 & 0 & 0.8986 & 1.0000 & 1.0000 & 0.9449 \\
      3 & 1 & 0.9094 & 1.0000 & 1.0000 & 0.9506 \\
      2 & 2 & 0.7558 & 1.0000 & 1.0000 & 0.8667 \\
      1 & 3 & 0.6191 & 1.0000 & 1.0000 & 0.7958 \\
      0 & 4 & 0.6941 & 1.0000 & 1.0000 & 0.8357 \\
      \bottomrule
    \end{tabular}%
  }
  \caption{Ablation over Standard vs.\ Rel-Aware Layers (total = 4).}
  \label{tab:ablation-layers}
\end{table}
\subsubsection{Layer Composition}
Table~\ref{tab:ablation-layers} shows that (i) 4 standard layers (0 Rel) achieve only 0.8986 Acc@1, indicating limited directional modeling; (ii) 3 standard + 1 Rel-Aware layers reach the best Acc@1 (0.9094) and MRR (0.9506), demonstrating that a single bearing-biased attention layer effectively complements temporal encoding; and (iii) allocating all 4 layers to Rel-Aware blocks drops Acc@1 to 0.6941, confirming that relational bias without temporal context harms performance. Thus, one Rel-Aware layer atop standard layers provides the optimal balance.

\subsubsection{POI Aggregation Granularity}
As shown in Table~\ref{tab:ablation-sectors}, \(S=8\) offers the best trade-off because it balances (i) \emph{directional aliasing} and \emph{quantization error}—too few sectors (e.g., \(S\in\{2,4\}\)) merge bearings that correspond to distinct movement intentions, diluting the mutual information between sector histograms and next-node choice; (ii) \emph{estimation variance}—with \(R=150\,\mathrm{m}\), increasing \(S\) amplifies normalization noise, which our \(\Delta\)-features further accentuate, so overly fine bins (e.g., \(S=16\)) become data-sparse and brittle. In short, \(S=8\) matches the bias–variance profile of the data and the capacity of our 3+1 layer stack, yielding the most reliable signal to the downstream Transformer.

\subsubsection{Radius Analysis}

Table~\ref{tab:coverage-radius} summarizes coverage metrics for radius \(R\in[50,250]\) m.  
Coverage and avg. POIs grow with \(R\), but marginal gain peaks at \(R=125\) m (0.77) and then declines; at \(R=150\) m we achieve high coverage (0.63) with moderate duplication (0.52) and still strong marginal gain (0.62).  
Thus, \(R=150\) m offers the best trade-off between context richness and feature compactness.

Overall, these ablations comprehensively validate the necessity and effectiveness of our semantic, geometric, and relational components, as well as our chosen aggregation parameters.

\section*{Conclusion}

We have presented a novel intersection-centric next-step prediction model that lifts the closed-world POI restriction by modeling trajectories on the full road-intersection graph and enriching each node with sector-wise POI-region projections. Our Rel-Aware LNN-Transformer merges dynamic temporal encoding and global directional bias within a compact, parameter-efficient architecture, yielding substantial gains over state-of-the-art baselines across diverse path lengths. Extensive tests confirm that the model maintains performance under realistic GPS jitter and semantic noise, validating its robustness in real-world conditions.

We restrict prediction to one-hop neighbors—the smallest topologically reachable set—enabling real-time inference and reflecting immediate decision points; extending our framework to k-hop candidate sets is a promising avenue for capturing longer-range planning behavior.

Furthermore, the lightweight design and efficient inference make our approach suitable for real-time deployment in navigation and traffic management systems, while the combined semantic and topological embeddings provide valuable insights for urban planners and mobility analysts.

\begin{table}[t]
  \centering
  \scriptsize

  \resizebox{\columnwidth}{!}{%
    \begin{tabular}{c|cccc}
      \toprule
      Sector Count & Acc@1 & Acc@3 & Acc@5 & MRR \\
      \midrule
      2  & 0.8734 & 0.9766 & 1.0000 & 0.9237 \\
      4  & 0.8818 & 0.9896 & 1.0000 & 0.9350 \\
      8  & 0.9094 & 1.0000  & 1.0000 & 0.9506 \\
     16  & 0.8704 & 0.9820 & 1.0000 & 0.9232 \\
      \bottomrule
    \end{tabular}%
  }
  \caption{Sector‐count Ablation (3 standard + 1 Rel-Aware layers).}
  \label{tab:ablation-sectors}
\end{table}

\begin{table}[t]
  \centering
  \small
  \setlength{\tabcolsep}{4pt}  % reduce column padding

  \resizebox{\columnwidth}{!}{%
    \begin{tabular}{rrrrr}
      \toprule
      \(\boldsymbol{R}\) (m) & \(\text{coverage (\%)}\) & \(\text{duplicate (\%)}\) & \(\text{avg\_nodes}\) & \(\text{marginal\_gain}\) \\
      \midrule
       50 & 0.2897 & 0.3665 & 1.5785 & 0.0000 \\
       75 & 0.3788 & 0.4222 & 1.7307 & 0.5855 \\
      100 & 0.4566 & 0.4562 & 1.8389 & 0.7189 \\
      125 & 0.5411 & 0.4867 & 1.9481 & 0.7739 \\
      150 & 0.6314 & 0.5226 & 2.0949 & 0.6150 \\
      175 & 0.7099 & 0.5633 & 2.2901 & 0.4022 \\
      200 & 0.7721 & 0.6060 & 2.5378 & 0.2510 \\
      225 & 0.8206 & 0.6474 & 2.8360 & 0.1628 \\
      250 & 0.8665 & 0.6818 & 3.1422 & 0.1498 \\
      \bottomrule
    \end{tabular}%
  }
  \caption{Overall coverage metrics as a function of sector radius \(R\).}
  \label{tab:coverage-radius}
\end{table}

\begin{itemize}
  \item \textbf{Multi-city generalization:} Evaluating transferability across diverse urban layouts in different cities and countries.  
  \item \textbf{Real-time sensory fusion:} Integrating live visual or environmental signals (e.g., camera feeds, weather data) for richer context.  
  \item \textbf{Crowd interaction modeling:} Extending from single trajectories to group-level dynamics, capturing how pedestrian interactions shape route choices.  
\end{itemize}

%\bibliographystyle{aaai_paper/aaai2026}
%\bibliography{aaai_paper/references}

\begin{thebibliography}{19}
\providecommand{\natexlab}[1]{#1}

\bibitem[{Cuttone, Lehmann, and González(2016)}]{CuttoneLehmannGonzalez2016}
Cuttone, A.; Lehmann, S.; and González, M.~C. 2016.
\newblock Understanding Predictability and Exploration in Human Mobility.
\newblock \emph{arXiv preprint arXiv:1608.01939}.
\newblock Version v1, 5 Aug 2016.

\bibitem[{Feng and et~al.(2018)}]{feng2018deepmove}
Feng, J.; and et~al. 2018.
\newblock DeepMove: Predicting Human Mobility with Attentional Recurrent Networks.
\newblock In \emph{WWW}.

\bibitem[{Grover and Leskovec(2016)}]{grover2016node2vec}
Grover, A.; and Leskovec, J. 2016.
\newblock {Node2Vec}: Scalable Feature Learning for Networks.
\newblock In \emph{Proceedings of the 22nd ACM SIGKDD International Conference on Knowledge Discovery and Data Mining}, 855--864.

\bibitem[{Guo and et~al.(2020)}]{guo2020arnn}
Guo, J.; and et~al. 2020.
\newblock {ARNN}: An Attentional Recurrent Neural Network for Personalized Next Location Recommendation.
\newblock In \emph{AAAI}.

\bibitem[{Hamilton, Ying, and Leskovec(2017)}]{hamilton2017graphsage}
Hamilton, W.~L.; Ying, R.; and Leskovec, J. 2017.
\newblock {GraphSAGE}: Inductive Representation Learning on Large Graphs.
\newblock In \emph{Advances in Neural Information Processing Systems}, volume~30, 1024--1034.

\bibitem[{Hasani et~al.(2022)Hasani, Amini, Rus, and Grosu}]{hasani2022cfc}
Hasani, R.; Amini, A.; Rus, D.; and Grosu, R. 2022.
\newblock Closed-form Continuous-time Neural Networks with Forgetting.
\newblock \emph{Nature Machine Intelligence}, 4(11): 1025--1036.

\bibitem[{Hasani et~al.(2020)Hasani, Lechner, Amini, Rus, and Grosu}]{hasani2020ltc}
Hasani, R.; Lechner, M.; Amini, A.; Rus, D.; and Grosu, R. 2020.
\newblock Liquid Time-Constant Networks.
\newblock In \emph{Proceedings of the Thirty-Fourth AAAI Conference on Artificial Intelligence}, 3319--3327.

\bibitem[{Hasani, Lechner, and Grosu(2023)}]{hasani2023scirobotics}
Hasani, R.; Lechner, M.; and Grosu, R. 2023.
\newblock Robust Vision-based Flight with Liquid Neural Networks.
\newblock \emph{Science Robotics}, 8(75).

\bibitem[{Hochreiter and Schmidhuber(1997)}]{hochreiter1997lstm}
Hochreiter, S.; and Schmidhuber, J. 1997.
\newblock Long Short-Term Memory.
\newblock \emph{Neural Computation}, 9(8): 1735--1780.

\bibitem[{Kong and Wu(2018)}]{kong2018hstlstm}
Kong, D.; and Wu, F. 2018.
\newblock HST-LSTM: A Hierarchical Spatial-Temporal Long-Short Term Memory Network for Location Prediction.
\newblock In \emph{Proceedings of the Twenty-Seventh International Joint Conference on Artificial Intelligence (IJCAI-18)}, 2341--2347.

\bibitem[{{Overpass Turbo}(2025)}]{overpass-turbo}
{Overpass Turbo}. 2025.
\newblock Overpass API Web Interface.
\newblock \url{https://overpass-turbo.eu/}.

\bibitem[{Xu, Huang, and Zou(2023)}]{TrajTrans}
Xu, S.; Huang, Q.; and Zou, Z. 2023.
\newblock Spatio‐Temporal Transformer Recommender: Next Location Recommendation with Attention Mechanism by Mining the Spatio‐Temporal Relationship between Visited Locations.
\newblock \emph{ISPRS International Journal of Geo‐Information}, 12(2): 79.

\bibitem[{Xu et~al.(2023)Xu, Suzumura, Yong, Hanai, Yang, Kanezashi, Jiang, and Fukushima}]{xu2023mobgt}
Xu, X.; Suzumura, T.; Yong, J.; Hanai, M.; Yang, C.; Kanezashi, H.; Jiang, R.; and Fukushima, S. 2023.
\newblock Revisiting Mobility Modeling with Graph: A Graph Transformer Model for Next Point-of-Interest Recommendation.
\newblock In \emph{Proceedings of the 31st ACM SIGSPATIAL International Conference on Advances in Geographic Information Systems (SIGSPATIAL '23)}, 94:1--94:10.

\bibitem[{Yang, Liu, and Zhao(2023)}]{yang2023getnext}
Yang, S.; Liu, J.; and Zhao, K. 2023.
\newblock GETNext: Trajectory Flow Map Enhanced Transformer for Next POI Recommendation.
\newblock \emph{arXiv preprint arXiv:2303.04741}.

\bibitem[{Zhao et~al.(2018)Zhao, Zhang, Liu, Yu, and Yang}]{zhao2018stlstm}
Zhao, J.; Zhang, W.; Liu, C.; Yu, H.; and Yang, K. 2018.
\newblock Spatio-Temporal LSTM with Trust Gates for 3D Human Action Recognition.
\newblock In \emph{European Conference on Computer Vision}, 816--833.

\bibitem[{Zhao et~al.(2019)Zhao, Zhu, Liu, Xu, Li, Zhuang, Sheng, and Zhou}]{zhao2019stgn}
Zhao, P.; Zhu, H.; Liu, Y.; Xu, J.; Li, Z.; Zhuang, F.; Sheng, V.~S.; and Zhou, X. 2019.
\newblock Where to Go Next: A Spatio-Temporal Gated Network for Next POI Recommendation.
\newblock In \emph{Proceedings of the Thirty-Third AAAI Conference on Artificial Intelligence (AAAI-19)}, 5877--5884.

\bibitem[{Zheng et~al.(2008)Zheng, Li, Chen, Xie, and Ma}]{zheng2008understanding}
Zheng, Y.; Li, Q.; Chen, Y.; Xie, X.; and Ma, W. 2008.
\newblock Understanding Mobility Based on GPS Data.
\newblock In \emph{Proceedings of the 10th International Conference on Ubiquitous Computing (UbiComp)}, 312--321. ACM.

\bibitem[{Zheng, Xie, and Ma(2010)}]{zheng2010geolife}
Zheng, Y.; Xie, X.; and Ma, W. 2010.
\newblock GeoLife: A Collaborative Social Networking Service among User, Location and Trajectory.
\newblock \emph{IEEE Data Engineering Bulletin}, 33(2): 32--40.

\bibitem[{Zheng et~al.(2009)Zheng, Zhang, Xie, and Ma}]{zheng2009mining}
Zheng, Y.; Zhang, L.; Xie, X.; and Ma, W. 2009.
\newblock Mining Interesting Locations and Travel Sequences from GPS Trajectories.
\newblock In \emph{Proceedings of the 18th International World Wide Web Conference (WWW)}, 791--800. ACM.

\end{thebibliography}

% \input{aaai_paper/ReproducibilityChecklist}
\end{document}